\newif\ifanon\anonfalse

\documentclass[a4paper]{article}

\usepackage{subcaption}
\usepackage{natbib}
\usepackage{tabularx}
\usepackage{times}
\usepackage{amssymb}
\usepackage{amstext}
\usepackage{amsmath}
\usepackage{amsthm}
\usepackage{stmaryrd}
\usepackage{todonotes}
\usepackage{enumitem}
\usepackage{multicol}
\usepackage{url}
\usepackage[capitalize]{cleveref}
\usepackage[verbose]{newunicodechar}
\newunicodechar{ }{{\,}}
\newunicodechar{¬}{\ensuremath{\neg}} % 
\newunicodechar{²}{^2}
\newunicodechar{³}{^3}
\newunicodechar{·}{\ensuremath{\cdot}}
\newunicodechar{¹}{^1}
\newunicodechar{×}{\ensuremath{\times}} % 
\newunicodechar{÷}{\ensuremath{\div}} % 
\newunicodechar{ʳ}{^r}
\newunicodechar{ˡ}{^l}
\newunicodechar{Γ}{\ensuremath{\Gamma}}   %
\newunicodechar{Δ}{\ensuremath{\Delta}} % 
\newunicodechar{Η}{\ensuremath{\textrm{H}}} % 
\newunicodechar{Θ}{\ensuremath{\Theta}} % 
\newunicodechar{Λ}{\ensuremath{\Lambda}} % 
\newunicodechar{Ξ}{\ensuremath{\Xi}} % 
\newunicodechar{Π}{\ensuremath{\Pi}}   %
\newunicodechar{Σ}{\ensuremath{\Sigma}} % 
\newunicodechar{Φ}{\ensuremath{\Phi}} % 
\newunicodechar{Ψ}{\ensuremath{\Psi}} % 
\newunicodechar{Ω}{\ensuremath{\Omega}} % 
\newunicodechar{α}{\ensuremath{\mathnormal{\alpha}}}
\newunicodechar{α}{\ensuremath{\mathnormal{\alpha}}} % 
\newunicodechar{β}{\ensuremath{\beta}} % 
\newunicodechar{γ}{\ensuremath{\mathnormal{\gamma}}} % 
\newunicodechar{δ}{\ensuremath{\mathnormal{\delta}}} % 
\newunicodechar{ε}{\ensuremath{\mathnormal{\varepsilon}}} % 
\newunicodechar{ζ}{\ensuremath{\mathnormal{\zeta}}} % 
\newunicodechar{η}{\ensuremath{\mathnormal{\eta}}} % 
\newunicodechar{θ}{\ensuremath{\mathnormal{\theta}}} % 
\newunicodechar{ι}{\ensuremath{\mathnormal{\iota}}} % 
\newunicodechar{κ}{\ensuremath{\mathnormal{\kappa}}} % 
\newunicodechar{λ}{\ensuremath{\mathnormal{\lambda}}} % 
\newunicodechar{μ}{\ensuremath{\mathnormal{\mu}}} % 
\newunicodechar{ν}{\ensuremath{\mathnormal{\mu}}} % 
\newunicodechar{ξ}{\ensuremath{\mathnormal{\xi}}} % 
\newunicodechar{π}{\ensuremath{\mathnormal{\pi}}}
\newunicodechar{π}{\ensuremath{\mathnormal{\pi}}} % 
\newunicodechar{ρ}{\ensuremath{\mathnormal{\rho}}} % 
\newunicodechar{σ}{\ensuremath{\mathnormal{\sigma}}} % 
\newunicodechar{τ}{\ensuremath{\mathnormal{\tau}}} % 
\newunicodechar{φ}{\ensuremath{\mathnormal{\varphi}}} % 
\newunicodechar{χ}{\ensuremath{\mathnormal{\chi}}} % 
\newunicodechar{ψ}{\ensuremath{\mathnormal{\psi}}} % 
\newunicodechar{ω}{\ensuremath{\mathnormal{\omega}}} %  
\newunicodechar{ϕ}{\ensuremath{\mathnormal{\phi}}} % 
\newunicodechar{ϵ}{\ensuremath{\mathnormal{\epsilon}}} % 
\newunicodechar{ᵏ}{^k}
\newunicodechar{ᵢ}{_i} % 
\newunicodechar{ }{{\,}}
\newunicodechar{′}{\ensuremath{^\prime}}  % 
\newunicodechar{″}{\ensuremath{^\second}} % 
\newunicodechar{‴}{\ensuremath{^\third}}  % 
\newunicodechar{ⁱ}{^i}
\newunicodechar{⁵}{\ensuremath{^5}}
\newunicodechar{⁺}{\ensuremath{^+}} %% 
\newunicodechar{⁺}{^+}
\newunicodechar{⁻}{\ensuremath{^-}} %% 
\newunicodechar{ⁿ}{^n}
\newunicodechar{₀}{\ensuremath{_0}} % 
\newunicodechar{₁}{\ensuremath{_1}}
\newunicodechar{₁}{_1}
\newunicodechar{₂}{\ensuremath{_2}}
\newunicodechar{₂}{_2}
\newunicodechar{₊}{\ensuremath{_+}} %% 
\newunicodechar{₋}{\ensuremath{_-}} %% 
\newunicodechar{ℂ}{\ensuremath{\mathbb{C}}} % 
\newunicodechar{ℕ}{\mathbb{N}} % 
\newunicodechar{ℝ}{\ensuremath{\mathbb{R}}} % 
\newunicodechar{ℤ}{\ensuremath{\mathbb{Z}}} % 
\newunicodechar{ℳ}{\mathcal{M}}
\newunicodechar{⅋}{\ensuremath{\parr}} % 
\newunicodechar{←}{\ensuremath{\leftarrow}} % 
\newunicodechar{↑}{\ensuremath{\uparrow}} % 
\newunicodechar{→}{\ensuremath{\rightarrow}} %
\newunicodechar{↔}{\ensuremath{\leftrightarrow}} % 
\newunicodechar{↖}{\nwarrow} %
\newunicodechar{↗}{\nearrow} %
\newunicodechar{↝}{\ensuremath{\leadsto}}
\newunicodechar{↦}{\ensuremath{\mapsto}}
\newunicodechar{⇆}{\ensuremath{\leftrightarrows}} % 
\newunicodechar{⇐}{\ensuremath{\Leftarrow}} % 
\newunicodechar{⇒}{\ensuremath{\Rightarrow}} % 
\newunicodechar{⇔}{\ensuremath{\Leftrightarrow}} % 
\newunicodechar{∀}{\ensuremath{\forall}}   %
\newunicodechar{∃}{\ensuremath{\exists}} % 
\newunicodechar{∅}{\ensuremath{\varnothing}} % 
\newunicodechar{∈}{\ensuremath{\in}}
\newunicodechar{∉}{\ensuremath{\not\in}} % 
\newunicodechar{∋}{\ensuremath{\ni}}  % 
\newunicodechar{∎}{\ensuremath{\qed}}
\newunicodechar{∏}{\prod}
\newunicodechar{∑}{\sum}
\newunicodechar{∗}{\ensuremath{\ast}} % 
\newunicodechar{∘}{\ensuremath{\circ}} % 
\newunicodechar{∙}{\ensuremath{\bullet}} % 
\newunicodechar{∙}{\ensuremath{\cdot}}
\newunicodechar{∞}{\ensuremath{\infty}} % 
\newunicodechar{∣}{\ensuremath{\mid}} % 
\newunicodechar{∧}{\wedge}% 
\newunicodechar{∨}{\vee}% 
\newunicodechar{∩}{\ensuremath{\cap}} % 
\newunicodechar{∪}{\ensuremath{\cup}} % 
\newunicodechar{∫}{\int}
\newunicodechar{∷}{::} % 
\newunicodechar{∼}{\ensuremath{\sim}} % 
\newunicodechar{≃}{\ensuremath{\simeq}} % 
\newunicodechar{≅}{\ensuremath{\cong}} % 
\newunicodechar{≈}{\ensuremath{\approx}} % 
\newunicodechar{≜}{\ensuremath{\stackrel{\scriptscriptstyle {\triangle}}{=}}} 
\newunicodechar{≜}{\triangleq}
\newunicodechar{≝}{\ensuremath{\stackrel{\scriptscriptstyle {\text{def}}}{=}}}
\newunicodechar{≟}{\ensuremath{\stackrel {_\text{\textbf{?}}}{\text{\textbf{=}}\negthickspace\negthickspace\text{\textbf{=}}}}}
\newunicodechar{≠}{\ensuremath{\neq}}% 
\newunicodechar{≡}{\ensuremath{\equiv}}% 
\newunicodechar{≤}{\ensuremath{\le}} % 
\newunicodechar{≥}{\ensuremath{\ge}} % 
\newunicodechar{⊂}{\ensuremath{\subset}} % 
\newunicodechar{⊃}{\ensuremath{\supset}} %  
\newunicodechar{⊆}{\ensuremath{\subseteq}} %  
\newunicodechar{⊇}{\ensuremath{\supseteq}} % 
\newunicodechar{⊎}{\ensuremath{\uplus}} %
\newunicodechar{⊑}{\ensuremath{\sqsubseteq}} % 
\newunicodechar{⊒}{\ensuremath{\sqsupseteq}} % 
\newunicodechar{⊓}{\ensuremath{\sqcap}} % 
\newunicodechar{⊔}{\ensuremath{\sqcup}} % 
\newunicodechar{⊕}{\ensuremath{\oplus}} % 
\newunicodechar{⊗}{\ensuremath{\otimes}} % 
\newunicodechar{⊢}{\ensuremath{\vdash}} % 
\newunicodechar{⊥}{\ensuremath{\bot}} % bottom
\newunicodechar{⊧}{\models} % 
\newunicodechar{⊨}{\models} % 
\newunicodechar{⊩}{\Vdash}
\newunicodechar{⊸}{\ensuremath{\multimap}} % 
\newunicodechar{⋁}{\ensuremath{\bigvee}}
\newunicodechar{⋃}{\ensuremath{\bigcup}} % 
\newunicodechar{⋄}{\ensuremath{\diamond}} % 
\newunicodechar{⋆}{\ensuremath{\star}} % 
\newunicodechar{⋮}{\ensuremath{\vdots}} % 
\newunicodechar{⋯}{\ensuremath{\cdots}} % 
\newunicodechar{─}{---}
\newunicodechar{□}{\ensuremath{\square}} % 
\newunicodechar{▹}{\ensuremath{\rhd}} % 
\newunicodechar{◃}{\triangleleft{}}
\newunicodechar{◇}{\ensuremath{\diamond}} % 
\newunicodechar{◽}{\ensuremath{\square}}
\newunicodechar{★}{\ensuremath{\star}}   %
\newunicodechar{♭}{\ensuremath{\flat}} % 
\newunicodechar{♯}{\ensuremath{\sharp}} % 
\newunicodechar{✓}{\ensuremath{\checkmark}} % 
\newunicodechar{⟂}{\ensuremath{^\bot}} % PERPENDICULAR 
\newunicodechar{⟦}{\ensuremath{\llbracket}}
\newunicodechar{⟧}{\ensuremath{\rrbracket}}
\newunicodechar{⟨}{\ensuremath{\langle}} % 
\newunicodechar{⟩}{\ensuremath{\rangle}} % 
\newunicodechar{⟶}{{\longrightarrow}}
\newunicodechar{⟹}{\ensuremath{\Longrightarrow}} % 
\newunicodechar{𝒟}{\ensuremath{\mathcal{D}}} % 
\newunicodechar{𝒢}{\ensuremath{\mathcal{G}}}
\newunicodechar{𝒦}{\ensuremath{\mathcal{K}}} % 
\newunicodechar{𝔸}{\ensuremath{\mathds{A}}} % 
\newunicodechar{𝔹}{\ensuremath{\mathds{B}}} % 
\newunicodechar{𝟙}{\ensuremath{\mathds{1}}} % 
\newunicodechar{̷}{\not} % 

\newlist{lingex}{enumerate}{3}
\setlist[lingex,1]{parsep=0pt,itemsep=1pt,label=(\arabic*),resume=lingexcount}
\newcommand\onelingex[1]{\begin{lingex}\item #1 \end{lingex}}

\newlist{subex}{enumerate}{3}
\setlist[subex,1]{parsep=0pt,itemsep=1pt,font=\bfseries,label=P\arabic*}
\newcommand\hyp{\item[H]}
\newcommand\answer{\item[A]}
\newcommand\failsf{\item[F]}
\newcommand\fracasex[2]{\begin{lingex}\item[(#1)] \begin{subex} #2 \end{subex} \end{lingex} }

\newcommand\constant[1]{\mathsf{#1}}

\begin{document}

\title{FraCaS: Temporal Analysis}
% \author{\ifanon Anonymous for review\else
%   \authorname{Jean-Philippe Bernardy} \authorname{Stergios Chatzikyriakidis}
%   \affiliation{University of Gothenburg}
%   \email{firstname.lastname@gu.se}\fi}

\author{\ifanon Anonymous for review\else
  {Jean-Philippe Bernardy}\footnote{University of Gothenburg} ~~ {Stergios Chatzikyriakidis}\footnote{University of Gothenburg}
}

\maketitle

% \keywords{Temporal Semantics, Type-Theory}

\abstract{In this paper, we propose an implementation of temporal semantics
  which is suitable for inference problems. This implementation
  translates syntax trees to logical formulas, suitable for
  consumption by the Coq proof assistant. We support several phenomena
  including: temporal references, temporal adverbs, aspectual classes
  and progressives.  We apply these semantics to the complete FraCaS testsuite.
  We obtain an accuracy of 81 percent overall and 73 percent for problems explicitly marked
  as related to temporal reference.
}

% \onecolumn \maketitle \normalsize \setcounter{footnote}{0} \vfill

\section{Introduction}
\label{sec:introduction}

The semantics of tense and aspect has been a long standing issue in
the study of formal semantics since the early days of Montague Grammar
and a number of different ideas have been put forth on the table to
deal with them throughout the years. Recent proposals include the
works of the following authors:
\citet{dowty:2012,prior:2003,steedman_productions_2000,higginbotham:2009,fernando:2015}.
The
semantics of tense and aspect have been also considered in the study of
Natural Language Inference (NLI). The various datasets
for NLI that have been proposed by the years contain examples that
have some implicit or explicit reliance on inferences related to tense
and aspect. One of the early datasets used to test logical approaches,
the FraCaS test suite~\citep{cooper:1996} contains a whole section
dedicated to temporal and aspectual inference (section 7 of the dataset). This part
of the FraCaS test suite has been difficult to tackle. That is,
so far, no computational system has been capable to deal with it in its entirety: when
authors report accuracy over the FraCaS test suite they
skip this section.
In fact, they also often skip the anaphora and ellipsis sections, the
exception being the system presented by
\citet{bernardy_type-theoretical_2017,bernardy_wide-coverage_2019},
which includes support for anaphora and ellipsis but still omit the
temporal section.\footnote{One can consider that \citet{maccartney:2007} have made a run against the whole test suite. However, they do not deal with multi-premise cases. Consequently only 36/75 cases in the temporal section are attempted. The general accuracy of the system is .59, and .61 for the temporal section. Our system, as shown \cref{tab:results}, presents considerable improvements in coverage and accuracy over that of \citeauthor{maccartney:2007}.}
In this paper, we take up the challenge of providing a
computationally viable account of tense and aspect to deal with the
  section 7 of the FraCaS test suite. Our account is not meant
to be a theoretically extensive account of tense and aspect, but rather an
account that is driven by the need to cover the test suite in a way
that is general enough to capture the test suite examples, \emph{while still covering the rest of the FraCaS test suite}.

The account is evaluated on the entailment properties of various
temporal and aspectual examples, as given by the test suite. As such,
we are not getting into the discussion of how tense and aspect might
affect grammaticality or infelicitousness of various sentences.
We assume that the sentences of the FraCaS suite are syntactically and
semantically correct, and strive to produce accurate logical
representations given that assumption. We further assume that the
entailment annotations of various problems are valid, and we use those
to evaluate the correctness of the logical representations of
sentences.

The paper is structured as follows: in \cref{sec:background}, we give
a brief summary of the computational frameworks whose various
subsystems rely on. In particular, the Grammatical Framework is used
to construct the syntactic parser, the Coq proof assistant checks all
the reasoning and a monad-based dynamic semantics deals with
Montague-style semantics, and references (anaphora).  We also provide
some brief remarks on temporal semantics. In \cref{sec:our-semantics},
we discuss the main aspects of the compositional semantics of our
system, using various examples from the FraCaS suite to illustrate its
effectiveness. In \cref{sec:results}, we evaluate how our system
performs with respect to the FraCaS suite. We ran the system across
the whole suite: our system is thus the first which is capable of
handling the complete FraCaS test suite.  Yet, we are interested in
particular in the performance on the temporal section. In
\cref{sec:conclusion}, we conclude and discuss avenues for future
work.

\section{Temporal-Semantics in a Logic-based NLI System}
\label{sec:background}

Our temporal analysis places itself in the context of a complete NLI
system -- which is why we can test it on the FraCaS suite. In this
section we give a brief overview of the phases of the system,
referring the reader to published work for details.

\paragraph{GF}
The first phase of the system, parsing, is taken care of by the 
Grammatical Framework (GF, \citet{ranta_grammatical_2004}), which is a
powerful parser generator for natural languages, based on
type-theoretical abstract grammars.  The present work leverages a
syntactic representation of the FraCaS test suite in GF abstract
syntax~\citep{Ljunglof:2012}. Thanks to this, we skip the parsing
phase and avoid any syntactic ambiguity.

For the purpose of this paper, the important feature of GF
syntax is that it aims at a balance of sufficient abstraction to provide a
semantically-relevant structure, but at the same time it embeds
sufficiently many syntactic features to be able to reconstruct
natural-language text. That is, the parse trees generally satisfy the
homomorphism requirement of \citet{montague_english_1970,montague_proper_1974}, and
we can focus on the translation of syntactic trees to logical forms.
Consequently, the system presented
here does not aim at textual natural language understanding, but rather
provides a testable, systematic formal semantics of temporal phenomena.
Example \ref{ex:nobel} shows an example abstract syntax tree and its
realisation in English.

\paragraph{Dynamic Semantics}
Parse trees are then processed by a dynamic semantic component.
Its role is essentially to support  (non-temporal) anaphora, using a
monadic-based dynamic semantics, generally following the state of the
art in this matter
\citep{unger_dynamic_2011,charlow_monadic_2015,charlow_modular_2017}.
Our particular implementation does not handle every case \ifanon
(citations anonymized) \else \citet{bernardy_computational_2020,bernardy_wide-coverage_2019} \fi
but non-temporal anaphoroi in the testsuite are generally resolved as
they should be: on the whole accuracy is not affected significantly by
issues in this subsystem.

As it is the case for other basic phenomena, there is not much
interaction between our treatment of time and non-temporal
anaphora. Critical exceptions are discussed in
\cref{sec:repeatable-achievements} and \cref{sec:interaction-with-anaphora}.

\paragraph{Montagovian Semantics}
Non-withstanding special support for anaphora, the core of the translation of syntax trees
to logical form follows a largely standard
montagovian semantics. In brief, sentences are interpreted as
propositions, verbs and noun-phrases as predicates.  We use
type-raising of noun-phrases, to support
quantifiers~\citep{montague_proper_1974}.

We support additionally the basic constructions and phenomena present
in the testsuite, including adjectives, adverbs, nouns, verbs,
anaphora, etc. The method is outlined by
\citet{montague_english_1970,montague_proper_1973}, but we direct the
reader to our previous work for details \ifanon (citations anonymized)
\else \citet{bernardy_type-theoretical_2017,bernardy_wide-coverage_2019} \fi, but the particular
treatment of such phenomena is essentially independent from our
treatment of time: in this paper we simply ignore these aspects beyond
the fact that they are handled correctly in most of the cases of the
FraCaS testsuite.

\paragraph{Inference using Coq}
Logical forms are then fed to the Coq
Coq interactive theorem prover (proof assistant). Coq is based on
the calculus of co-inductive constructions~\citep{werner_une_1994} We
do not use any co-induction (or even induction) in this paper, relying
on the pure lambda-calculus inner core of Coq.  Coq is a very powerful
reasoning engine that makes it fit for implementing natural language
semantics.  Coq also supports  dependent typing and 
 subtyping. Both concepts are instrumental in expressing NL
semantics~\citep{chatzikyriakidis_natural_2014}.  Besides, on a more
practical side, it works well for the the task of NLI, when the latter
is formalised as a theorem proving task: its many tactics mean that
many tasks in theorem proving are trivialised. In particular, all
problems of time-intervals inclusion, which occur in every temporal
problems, are solved with Coq's linear arithmetic tactic.

\section{Our Treatment of Time}
\label{sec:our-semantics}

In montagovian semantics, (intransitive) verbs are one-place
predicates; in types, they are functions from entities to propositions
($e \to t$).  Our basic approach is to generalise the interpretation of 
verbs, so that it takes two
additional time parameters, one corresponding to the starting time
of the action and one corresponding to its stopping time ($(e × time × time) \to t$). For example,
if John walked between $t_0$ and $t_1$, we would have:
$walk(john,t_0,t_1)$. From now on we will call an interval of time
points $[t_0,t_1]$ a timespan. Every timespan $[t_0,t_1]$ has the
property $t_0 ≤ t_1$ (it starts no later than it stops). (We are thus
using a simple Newtonian model of time, corresponding to a layman
intuition of a linear constant flow of time.)

In principle, common nouns and adjectives should undergo the same
procedure. For simplicity we will however only consider verbs from now
on. (In fact, even in our implementation we chose not to extend nouns
nor adjectives with timespan parameters. This choice limits the
increase in complexity of the formulas compared to non-temporal
semantics, at the expense of inaccuracy for a couple of problems in
the FraCaS test suite: problems 271 and 272 use a an adjective as a
copula which is subject to temporal reasoning.)

\paragraph{Temporal Context}

We adjust the montagovian semantics so that the interpretation of
every category (propositions, verb phrases, etc.) takes a
\emph{temporal context} as an additional parameter, which serves as a time
reference for the interpretation of all time-dependent semantics
within the phrase. (While some categories do not need this temporal context, we pass it everywhere for consistency.)  This context propagates through the compositional
interpretation down to lexical items with atomic representation
(verbs). By default, every interpretation passes the temporal context
down to its components without changing it. However some key elements
will act on it on nontrivial ways, which we proceed to detail below.

This temporal context is an \emph{optional} timespan. That is, it can
be a timespan or an explicitly unspecified context.
\newcommand\nospan{\ensuremath{-}}
\newcommand\anyspan{\ensuremath{\mathbf{*}}} The timespan in the context is
optional because, in certain situations, the semantics is different
depending on whether a timespan has been specified externally or not,
as we explain below. A non-present timespan will be represented as
\nospan{}. If a semantic function does not depend on the temporal
context at all, we will write \anyspan{} instead.

\paragraph{Tenses}

\newcommand\varid[1]{\mathnormal{#1}}
The principal non-trivial manipulators of timespans are tense
markers. In our syntax, inherited from GF, tenses are represented
syntactically as an attribute of clauses. An illustration of a
past-tense clause and its interpretation follows in
Example \ref{ex:nobel}. Notice in particular the $\varid{past}$ argument to
the $\varid{useCl}$ constructor.

\onelingex{A scandinavian won the nobel prize.\label{ex:nobel}
\\
{\small\begin{multline*}
\varid{useCl}\, \varid{past}\, \varid{pPos}\\ (\varid{predVP}\, (\varid{detCN}\, (\varid{detQuant}\, \varid{indefArt}\, \varid{numSg})\\ \varid{scandinavian\_CN})\\ (\varid{complSlash}\, (\varid{slashV2a}\, \varid{win\_V2})\\ (\varid{detCN}\, (\varid{detQuant}\, \varid{indefArt}\, \varid{numSg})\\ \varid{nobel\_prize\_CN})))
\end{multline*}
}}
In our semantics we deal only with present and past tenses (simple and
continuous). Indeed we find that FraCas does not exercise additional
specific tenses. (When a more complicated tense is used, the
additional information is also carried by adverbs or adverbial
phrases, in a more specific way). While we believe that many other
tenses can be captured under the same general framework, we leave a
detailed study to further work.

Even though we discuss a refinement to handle the past continuous
at the end of this section, the procedure to handle tense annotations
is as follows:
\begin{itemize}
\item If the tense is the past, and the temporal context is
  unspecified, then we locally quantify over a time interval
  $[t_0,t_1]$, such that $t_1 < now$, where $now$ is a logical
  constant representing the current timepoint. The temporal context
  then becomes this interval.
\item If the tense is the present and the temporal context is
  unspecified, then the temporal context becomes the simple
  $(now,now)$ interval.
\item If the temporal context is specified (for example due to the presence of an adverb or an
  adverbial clause, such as ``before James swam''), then the tense does
  not create a new interval, but it may constrain it. Typically, a
  past tense adds the constraint that the temporal context ends before
  the timepoint $now$.
\end{itemize}

\paragraph{Temporal Adverbs}

The other single most important source of interesting timespans are
adverbs. Most of the temporal adverbs fall in either of the following
categories:

\begin{enumerate}[align=left,font=\itshape]
\item[exact] For such adverbs, an exact interval is
  provided. In fact, such adverbs typically specify a single point in
  time (so the start and the end of the interval coincide).

  \[⟦\text{at 5 pm, \(s\)}⟧(\anyspan) = ⟦s⟧(5pm,5pm)\]

\item[existentially quantifying] The majority of temporal adverbs
  existentially quantify over a timespan. Examples include ``since
  1991'', ``in 1996'', ``for two years'', etc. The common theme is to
  introduce the interval and then restrict its bounds or its duration
  in some way. Sometimes the restriction is an equality, as in ``for
  exactly two hours''. In the following example we show the inclusion
  constraint, for ``in 1992''.

  \begin{multline*}
  ⟦\text{in 1992, \(s\)}⟧(\anyspan) = \\ ∃t_1,t_2.  [t_1,t_2] ⊆ 1992, ⟦s⟧(t_1,t_2)
\end{multline*}
In the FraCaS test suite, we normally do not find several
time-modifying adverbs modifying a single verb phrase. Indeed, sentences such
as ``in 1992, in 1991 john wrote a novel'' are infelicitous. This
justifies ignoring the input timespan in the above interpretation --
we are in particular not interested in modelling felicity with our
semantics, only giving an accurate semantics when the input is
felicitous.
\item[universally quantifying] A few adverbs introduce intervals via a
  universal quantification (sometimes with a constraint). Examples
  include ``always'' and ``never''.

  If there is no explicit time context, then ``always'' has no
  constraint on the interval, otherwise the quantified interval must
  be included in it:
  \begin{multline*}
  ⟦\text{always \(s\)}⟧(t_0,t_1) = \\ ∀t_0',t_1'. [t_0',t_1'] ⊆ [t_0,t_1], ⟦s⟧(t_0',t_1')
\end{multline*}
Note that here we \emph{do} use the input interval, resulting in a correct interpretation for
phrases such as ``In 1994, Itel was always on time.'' .

% A specifically interesting case found in FraCas is the adverbial phrase ``every month'', which we modeled
% \begin{equation}
%   ⟦\text{every month s}⟧(t_0,t_1) = ∀t. [t,t+1month] ⊆ [t_0,t_1], ⟦s⟧(t,t+1month)
% \end{equation}
\end{enumerate}

\paragraph{Aside: aspectual classes in the literature}
In this paper we borrow several notions from classical temporal
semantics such as ``stative'', ``achievement'', ``activity'', etc.,
even though our definitions do not perfectly match the classical
ones. We explain our precise meaning for these terms in the body of the
paper. Nevertheless, we refer the reader to \citet{steedman:2000} for
an extensive review of formal temporal semantics.

For the \textit{cognoscenti}, we can already point out some
differences in terminology: we use the term activity as a general term
which encompasses the three classical notions of activites,
achievements and accomplishments. Indeed, insofar as the test suite is
concerned, we find that these three categories can be collapsed into a
single one (they are subject to \cref{eq:unicity}).  As such, we think
it is a good classification, given that it generally affords the
correct inferences for the testsuite. (In this paper, we always assume
that the problems in the FraCaS testsuite are correctly annotated.)

\paragraph{Time references and aspectual classes}

A common theme in the testsuite is the reference to previous
occurrences of an event:

\fracasex{262}{
\item	Smith left after Jones left.
\item	Jones left after Anderson left.
\hyp 	Did Smith leave after Anderson left?
}

To be able to conclude that there is entailment, as the testsuite expects, we have to make sure
that the two occurrences of ``Jones left'' (in \textbf{P1} and \textbf{P2}) refer to the
same time intervals.
For this purpose we postulate \emph{unicity of action} for certain time-dependent
propositions:

\begin{multline}
  \constant{unicity}_P : P (t_1,t_2) → P (t_3,t_4) → \\ (t_1 = t_3) ∧ (t_2 = t_4)
  \label{eq:unicity}
\end{multline}

Unicity of action holds only if the aspectual class of the proposition
$P$ is \emph{activity}~\citep{steedman_productions_2000} (which, for our
purposes, includes \emph{achievements} and \emph{accomplishments}).

(The difference between activity and accomplishments on the one hand and
achievement on the other hand is that for the latter, time intervals
can be assumed to be of nil duration. In reality, this is an
oversimplification as achievements are usually of short duration, but
not nil. However, this plays little role in our
analysis. As far as we can tell the FraCaS test suite does exercise temporal semantics to
such a level of precision.)

Unicity of action plays the role of event coreference in
(neo-)Davidsonian accounts \citep{parsons1990events}. It is also a fine-grained
principle, allowing coreference to take into account certain arguments
when referencing. As we detail below, taking arguments into account
yields is critical to handle repeatability of achievements.

Unicity of action appears to be a non-logical principle. Indeed, it is
quite possible that ``Jones left'' several times. However, it seems
that this principle is never contradicted by the testsuite. As such,
even though unicity of action is only a pragmatic rule, it can be
taken as a valid one \emph{by default}: it is only when we have a
sufficiently constrained situation that one should reject it. Consider
the following discourse:
\begin{enumerate}[parsep=0pt,itemsep=1pt,label=(\arabic*)]
\item Smith left at 1pm.
\item Smith went to his appointment with the lawyer.
\item Smith left at 4pm.
\end{enumerate}
One would normally not say that there is contradiction. However if the
middle sentence were not present, a contradiction should be
flagged. We leave such discourse analysis as future work, and simply
apply unicity of action everywhere: it is valid uniformly in the
FraCaS test suite for activity aspect classes.

\paragraph{Statives}
\textit{A contrario}, if \(P\) is \emph{stative}, then we get a time-interval subsumption property:

\begin{multline*}
\constant {subsumption}_P :\\ [t_3,t_4] ⊆ [t_1,t_2] → P (t_1,t_2) → P (t_3,t_4)
\end{multline*}
This principle is used to reason about problem (314), below:

\fracasex{314}
{\item	Smith arrived in Paris on the 5th of May, 1995.
\item	Today is the 15th of May, 1995.
\item	She is still in Paris.
\hyp 	Smith was in Paris on the 7th of May, 1995.
}

Indeed, from \textbf{P3} we get that Smith was in Paris between May
5th and May 15th. Because ``being in Paris'' is stative, we also get that
Smith was in Paris in any sub-interval. Contrary to unicity of action,
subsumption is always valid.

\paragraph{Class-modifying adverbs}
It should be noted that some adverbs can locally disable the
application of $\constant {subsumption}$. For example, problem 299
features the sentence ``Smith lived in Birmingham for exactly a
year''.  Even though ``live'' is normally stative, one can no longer
apply subsumption in the context of ``exactly a year'' --- this can be
done by propagating another context flag in the montagovian semantics
(in addition to the temporal context).

\paragraph{(Un)repeatable Achievements}
\label{sec:repeatable-achievements}
The principle of using unicity of action interacts well with the usual
interpretation of existential quantifiers (and anaphora).  Indeed, using it, we
can refute problem (279), as expected by the testsuite:

\fracasex{279}{
\item	Smith wrote a novel in 1991.
\hyp 	 Smith wrote it in 1992.
}
Indeed, following our account, the above (contradictory) inference problem is to be
interpreted as

\begin{equation}
  \label{eq:smith-novel-1}
\begin{array}{l}
∀x. novel(x) ∧ \\
∃t_1,t_2. [t_1,t_2] ⊆ 1991 ∧ write(smith,x,t_1,t_2) ∧ \\
∃t_3,t_4. [t_3,t_4] ⊆ 1992 ∧ write(smith,x,t_3,t_4) \\
⟶ ⊥ \\
\end{array}
\end{equation}
Note here that the scope for the existential is extended beyond the
scope of \textbf{P1}, and its polarity switched (to universal). This extension can follow the account of
\citet{unger_dynamic_2011}, and our implemented analysis of
anaphora\ifanon (citation anonymized)
\else\citep{bernardy_computational_2018,bernardy_wide-coverage_2019}\fi.

Thanks to the unicity of action of $write(smith,x,...)$ (the subject
and direct object are fixed) we find \([t_1,t_2] = [t_3,t_4]\), and
due to the years 1991 and 1992 being disjoint we obtain contradiction.

However, the testsuite instructs that we should \emph{not} be able to refute
problem (280), with the justification that ``wrote a novel'' is a repeatable
accomplishment:

\fracasex{280}{
\item	Smith wrote a novel in 1991.
\hyp 	Smith wrote a novel in 1992.
}
Here our interpretation is:

\[\begin{array}{l}
(∃x. novel(x) ∧ \\
∃t_1,t_2. [t_1,t_2] ⊆ 1991 ∧ write(smith,x,t_1,t_2)) ∧ \\
(∃y. novel(y) ∧ \\
∃t_3,t_4. [t_3,t_4] ⊆ 1992 ∧ write(smith,y,t_3,t_4)) \\
⟶ ⊥ \\
\end{array}
\]
Our analysis does not need to treat this last case specially. Indeed,
even if $write(smith,x,.,.)$ is an activity and thus subject to
unicity of action, in (280), $x$ is quantified existentially; we have
two \emph{different} actions: $write(smith,x,t_1,t_2)$ and
$write(smith,y,t_3,t_4)$, because $x \neq y$, and thus we cannot
deduce equality of the intervals $t_1,t_2$ and $t_3,t_4$. In turn, the
hypothesis cannot be refuted.

\paragraph{Action-modification Verbs}

The final class of lexemes carrying a temporal-dependent semantics are
verbs taking a proposition as argument, like ``finish'', ``start'',
etc. These verbs modify the temporal context in non-trivial
ways. Consider for example ``finish to ...''. The timespan of the
argument of ``finish'' should end within the timespan of the finishing
action:

\begin{multline*}
⟦\text{finish to \(s\)}⟧(t_0,t_1) = \\ ∃(t_0',t_1'). t_1' ∈ [t_0,t_1] ∧ ⟦s⟧(t_0',t_1')
\end{multline*}

\paragraph{Progressive Aspect}

We treat verbs in the progressive form as different semantically from
the non-progressive form. For example, ``John was writing a book'' is
encoded as
$∃(t_1, t_2). t_1≤ t_2, t_2 ≤ now, PROG\_write(John,book,t_1,t_2)$,
while ``John wrote a book'' is encoded as
$∃(t_1, t_2). t_1≤ t_2, t_2 ≤ now, write(John,book,t_1,t_2)$. This distinction is
necessary because in our analysis the progressive form ($PROG\_write$) is subject to
$\constant{subsumption}$. That is, if John is writing in the interval
$[t_1,t_2]$ then he is writing in any sub-interval of
$[t_1,t_2]$. This interpretation corresponds to the idea the the action takes place continuously over the whole interval.
However, the same cannot be said of the non-continuous form ($write$): the
end-points of the interval indicate the time needed to complete the
achievement. (For example, ``John wrote a book in 1993'' neither
entails ``John wrote a book in January 1993'' nor ``John wrote a book
in December 1993''.) (In fact, \emph{write}, in the non-progressive
from, is on the contrary subject to $\constant{unicity}$.) Finally, we
also have $write(x,y,t_1,t_2) → PROG\_write(x,y,t_1,t_2)$. That is,
the achievement (or \emph{activity} in our terminology) variant implies the stative variant, for the
same interval.  Consequently we get the entailment from ``John wrote a
book in 1993'' to ``John was writing a book in 1993'', but not the other way around.

We note however that this interpretation differs only slightly from
the usual accounts of the progressive in the
literature. \Citet{ogihara2007tense} summarises the position of
\citet{bennett1978toward} as follows: a progressive sentence is true
at an interval $[t_0,t_1]$ iff there is an interval $[t'_0,t'_1]$ such
that $[t_0,t_1]$ is a non-final subinterval of $[t'_0,t'_1]$ and the
progressive sentence is true at $[t'_0,t'_1]$. This is very similar to
our approach ($\constant{subsumption}$ for the progressive form only),
but there is a difference regarding final intervals. Yet in our view
this difference is hard to justify: we cannot see why ``John was
writing a book in 1993'' entails that he was writing it January,
February, etc. but not in December.

\Citet{ogihara2007tense} argues that this simple account of the
progressive fails to reject a sentence such as ``Lee is resembling
Terri.'' while ``Lee is walking'' is acceptable. We argue instead that
the latter should be rejected for pragmatic reasons. Indeed, when a
predicate holds for a very long interval, one typically uses the
simple present tense in English. Therefore the continuous form
pragmatically implies that the predicate holds for a limited interval.
But, without further context, the predicate ``resemble Terri'' does
not vary over time (while ``walk'' generally does). Therefore the
continuous form ``Lee is resembling Terri'' is confusing: one implies
a limited interval, but the semantics of resembling normally yield an unlimited interval.
Because we do not account for pragmatics, we prefer to retain the
simplest account based on the subinterval property (which we call
$\constant{subsumption}$ here).

Finally we stress that not all verbs are subject to the
stative/achievement distinction induced by the progressive. For
example, the phrases ``John ran'' and ``John was running'' appear to
be logically equivalent, for entailment purposes.

\section{Worked out example}
To give a sense of the additional details necessary to deal with the
precision demanded by a proof-assistant such as Coq we show how
problem (279) is worked out in full details.

We start with input trees in GF format, given by
\citet{Ljunglof:2012}. They can be rendered as follows:

\begin{small}
\begin{verbatim}
s_279_1_p=
sentence
(useCl past pPos
 (predVP
   (usePN (lexemePN "smith_PN"))
   (advVP
     (complSlash
      (slashV2a (lexemeV2 "write_V2")))
     (detCN (detQuant indefArt numSg)
       (useN (lexemeN "novel_N"))))
       (lexemeAdv "in_1991_Adv")))
s_279_3_h=
sentence
(useCl past pPos
 (predVP (usePN
   (lexemePN "smith_PN"))
   (advVP
     (complSlash
      (slashV2a (lexemeV2 "write_V2"))
      (usePron it_Pron))
     (lexemeAdv "in_1992_Adv"))))
\end{verbatim}
\end{small}

Of particular note is the use of the pronoun ``it'', and the fact that
adverbials expressions such that ``in 1992'' are lexicalized.  We also
follow the GF convention to postfix lexical items with the name of
their category. Most of the other categories follow usual naming
conventions. We remind the reader that ``slash'' categories are used
to swap the order of arguments (compared to non-slashed categories of
similar names).

Our dynamic and temporal semantics gives the following interpretation
for \verb!s_279_1_p! implies \verb!s_279_3_h!.
{\small
\begin{verbatim}
FORALL (fun a=>novel_N a)
(fun a=>(exists (b: Time),
((exists (c: Time),
(IS_INTERVAL Date_19910101 b /\
 IS_INTERVAL c Date_19911231 /\
 IS_INTERVAL b c /\
 appTime b c (write_V2 a)
   (PN2object smith_PN))))) ->
Not (exists (f: Time),
((exists (g: Time),
 (IS_INTERVAL Date_19920101 f /\
  IS_INTERVAL g Date_19921231 /\
  IS_INTERVAL f g /\
  appTime f g (write_V2 a) 
    (PN2object smith_PN)))))).
\end{verbatim}
}
In the above, one should remark the top-level quantification over the
novel (as explained in \cref{sec:our-semantics}), the quantification
over time intervals as individual timepoints, and the use of custom
operators for several constructions (\verb!FORALL!, \verb!Not!, \verb!IS_INTERVAL!,
\verb!appTime!). This use of custom operators is useful for several
generalisations (for example, we have quantifiers such as \verb!MOST! in
addition to \verb!FORALL! --- see \ifanon (citation anonymized)
\else \citet{bernardy_type-theoretical_2017} \fi)

Unfolding the definitions for these operators yield the following
proposition:
{\small
\begin{verbatim}
forall x : object,
novel_N x ->
(exists b c : Z,
   Date_19910101 <= b /\
   c <= Date_19911231 /\
   b <= c /\ write_V2 x SMITH b c) ->
(exists f g : Z,
   Date_19920101 <= f /\
   g <= Date_19921231 /\
   f <= g /\ write_V2 x SMITH f g) ->
False
\end{verbatim}
}This is very close to our idealised representation of the problem
\cref{eq:smith-novel-1}. One difference is the use of abstract Coq
integers for timepoints. Using a discrete time allows us to use
predefined Coq tactics. The discrete nature of integers does not
interfere with the reasoning.

Finally, we can show a Coq proof for the above proposition:
{\small
\begin{verbatim}
Theorem  problem279 : Problem279aFalse.
cbv.
intros novel isSmithsNovel P1 H.
destruct P1 as
   [t0 [t1 [ct1 [ct2 [ct3 P1]]]]].
destruct H as
   [u0 [u1 [cu1 [cu2 [cu3 H]]]]].
specialize writeUnique
  with (x := novel)(y := SMITH) as A.
unfold UniqueActivity in A.
specialize (A _ _ _ _ P1 H) as B.
lia.
Qed.
\end{verbatim}
}

The intros and destruct tactics serve bookkeeping purposes. The
critical part is the use of the \texttt{writeUnique} axiom, which
witnesses the aspectual class of the predicate \texttt{write\_V2}.
The proof is completed by the use of the \texttt{lia} tactic, which is
embeds a decision procedure for linear arithmetic
problems. Fortunately, \texttt{lia} can take care of all the problems
which arise in the FraCaS testsuite.

\begin{figure*}
\begin{subfigure}{0.47\textwidth}

\fracasex{252}{\answer yes
\item	Since 1992 ITEL has been in Birmingham.
\item	It is now 1996.
\hyp    Itel was in Birmingham in 1993.
} \fracasex{260	}{\answer yes
\item	Yesterday APCOM signed the contract.
\item	Today is Saturday, July 14th.
\hyp    APCOM signed the contract Friday, 13th.

} \fracasex{262	}{\answer yes
\item	Smith left after Jones left.
\item	Jones left after Anderson left.
\hyp    Smith left after Anderson left.

} \fracasex{312	}{\answer yes
\item	ITEL always delivers reports late.
\item	In 1993 ITEL delivered reports.
\hyp    ITEL delivered reports late in 1993.

} \fracasex{320	}{\answer yes
\item	When Jones got his job at the CIA, he knew that he would never be allowed to write his memoirs.
\hyp    It is the case that Jones is not and will never be allowed to write his memoirs.
}

  \caption{Selected correctly handled examples}
  \label{fig:correct-examples}
\end{subfigure}
\hfill
\begin{subfigure}{0.47\textwidth}
{
} \fracasex{259	}{\answer yes
\item	The conference started on July 4th, 1994.
\item	It lasted 2 days.
\hyp    The conference was over on July 8th, 1994.
\failsf Incorrect lexical semantics for ``last $n$ days''.
} \fracasex{294	}{\answer yes
\item	Smith was running his own business in two years.
\hyp    Smith ran his own business.
\failsf Parsing error for ``in two years''
} \fracasex{317	}{\answer yes
\item	Every representative has read this report.
\item	No two representatives have read it at the same time.
\item	No representative took less than half a day to read the report.
\item	There are sixteen representatives.
\hyp    It took the representatives more than a week to read the report.
\failsf Counting semantics not implemented
% } \fracasex{318	}{\answer no
% \item	While Jones was updating the program, Mary came in and told him about the board meeting.
% \item	She finished before he did.
% \hyp    Mary's story lasted as long as Jones's updating the program.
% \failsf Incorrect interaction with anaphora
}
  \caption{Selected incorrectly handled examples.
  We list below a number of selected examples from the temporal section
of the FraCaS testsuite where our semantics fails to list the correct
inference. We list the reason why it fails under the item \textbf{F}.
}
  \label{fig:incorrect-examples}
\end{subfigure}
% \caption{Selected examples.}
\end{figure*}

\section{Results and Evaluation}
\label{sec:results}
Our target is the FraCaS testsuite, which aims at covering a wide
range of common natural-language phenomena. 
The suite is structured according to the semantic phenomena involved
in the inference process for each example, and contains nine sections:
Quantifiers, Plurals, Anaphora, Ellipsis, Adjectives, Comparatives,
Temporal, Verbs and Attitudes.  The system described here focuses on
the Temporal section. However, it also supports the other eight
sections.  To our knowledge this is the first system which attempts to
target the temporal section in full. But in fact, our system even provides
support for all the other sections. Thus, a couple of decades after its
formulation, we propose a first attempt at covering the whole suite.
As such, there it is no other system to compare our system with, in
all aspects.  We can however compare with systems which target parts
of the FraCaS testsuite, as shown in \cref{tab:results}.

\providecommand\ncases[1]{{\ensuremath{^{#1}}}}
\begin{table}[hbt]
  \centering
  \small
\begin{tabularx}{\columnwidth}{Xr@{\,\,}r@{\,\,}r@{\,\,}r@{\,\,}r@{\,\,}r@{\,\,}r}
Section      & {\kern -2em} \#FraCaS
                          & This        & FC2         & FC & MINE & Nut  & LP  \\ \hline
Quantifiers  & 75         & .93        & .96         & .96    & .77  & .53  & .93  \\
             &            & \ncases{74}& \ncases{74} &        &      &      &     \ncases{44} \\
Plurals      & 33         & .79        & .82         & .76    & .67  & .52  & .73 \\
             &            &            &             &     &   &   & \ncases{24} \\
Anaphora     & 28         & .79        & .86         &   -    & -    & -    &  -       \\
Ellipsis     & 52         & .81        & .87         &   -    & -    & -    &  -       \\
Adjectives   & 22         & .95        & .95         & .95    & .68  & .32  & .73 \\
             &            & \ncases{20}&  \ncases{20}&     &   &   &  \ncases{12} \\
Comparatives & 31         & .65        & .87         & .56    & .48  & .45  &  -       \\
Temporal     & 75         & .73        &  -          &   -    &   -  &  -   &  -       \\
Verbs        & 8          & .75        & .75         &   -    & -    & -    &  -       \\
Attitudes    & 13         & .85        & .92         & .85    & .77  & .46  & .92  \\ 
             &            &            &             &        &      &      & \ncases {9}  \\ \hline
Total        & 337        & .81        & .89         & .83    & .69  & .50  & .85  \\
             &            & \ncases{329}& \ncases{259}& \ncases{174}  & \ncases{174}& \ncases{174}& \ncases{89}
  \end{tabularx}
  \caption{Accuracy of our system compared to others.
    ``This" refers to the approach presented in this paper. When a
    system does not handle the nominal number of test cases (shown in
    the second column), the actual number of test cases attempted is
    shown below the accuracy figure, in smaller font.  ``FC''
    refers to the work of \citet{bernardy_type-theoretical_2017}, and ``FC2'' its followup~\citep{bernardy_wide-coverage_2019}. ``MINE" refers
    to the approach of \citet{Mineshima:2015}, ``NUT" to the CCG
    system that utilises the first-order automated theorem prover
    \textit{nutcracker} \cite{bos:2008}, and ``Langpro"
    to the system presented by \citet{abzianidze_tableau_2015}. A dash
    indicates that no attempt was made for the section. }
  \label{tab:results}
\end{table}
Additionally, to provide the reader with a better sense of what the
complete system is capable and not capable of, we list a number of
correctly and incorrectly handled examples in \cref{fig:correct-examples} and \cref{fig:incorrect-examples}.
\paragraph{Interaction with anaphora}
\label{sec:interaction-with-anaphora}
One reason explaining the lower performance of our system on some
sections of the testsuite is that our interpretation of time interacts
imperfectly with anaphora and ellipsis. Consider the following example:

\fracasex{232}
{\item	ITEL won more orders than APCOM did.
\item	APCOM won ten orders.
\hyp 	ITEL won at least eleven orders.
}

In the first premise, our system essentially resolves the ellipsis to
get the following reading: ``ITEL won \(X\) orders and APCOM won \(Y\) orders
and $X > Y$.''. One would need each of the verb phrases ``won \(X\) orders''
and ``won $Y$ orders'' to introduce their own timespans with existential
quantifiers. However, the organisation of our system is such that the
existentials are introduced before the ellipsis is
expanded. Consequently we get a wrong interpretation and the inference
cannot be made.

\section{Conclusions and Future Work}
\label{sec:conclusion}

We have presented a first attempt for a computational approach dealing
with the temporal section of the FraCaS test suite. To do this, we
have provided a new, simplified taxonomy of aspectual classes for verb
phrases, guided by the applicability of the unicity of action and
temporal subsumption properties. While part of this simplification is
accidental (conflation of activity and accomplishment), we find that
other parts (the automatic distinction between repeatable and
unrepeatable achievements) constitute theoretical improvements.

Besides inference, formal interpretation of tense is found in
natural-language interfaces to databases. Of note is the work of
\citet{androutsopoulos1998time}, which handles many of the time-aware
adverbial clauses that we address. However, we cover many more
 logical aspects of inference, such as coreference via
unity of action and interaction with quantifiers.

\cite{bernardy_wide-coverage_2019} presented a
logical system for handling 8 of the 9 sections of the FRACAS test
suite, but excluded section 7, suggesting that it requires many
examples that need an \textit{ad hoc} treatment. Here, we take up this
challenge and show that a system similar to theirs can be extended to
cover the remaining section of the test suite.
We use the same combination of a number of well-studied tools: type
theory, parsing using the Grammatical Framework (GF), Monadic Dynamic
Semantics and proof assistant technology (Coq). The system achieves an
accuracy of 0.73 on the Temporal Section and 0.81 overall.
One of the
things to be looked at is fixing the issues associated with parts of
the test suite that ``broke'' when the temporal analysis was
introduced. Some of these have been already mentioned: interaction of
the temporal variables with anaphora.

Another extension of this work is to reflect more temporal semantic
inference properties in an extended test suite. Indeed, there as properties which are
not captured in the FraCaS test suite, such as fine-grained examples of lexical and
grammatical aspect, as well as the interaction between those two, for example
cases where one needs to actually distinguish between achievements and
accomplishments on the basis of their inferential properties:

\fracasex{extra1}{
	\item	John found his keys.  
	\hyp 	John was finding his keys  (UNK).
}

\fracasex{extra2}{
	\item	John wrote a book.  
	\hyp 	John was writing a book  (YES).
}

In the first of the two examples involving an achievement verb, the
inference is UNK, since there is no guarantee that the action is
non-instantaneous. To the contrary, for accomplishment verbs, the
inference follows.

Further cases to be included in an extended FraCaS future suite
involve examples where the interaction between different tenses needs
to be captured:

\fracasex{extra3}{
	\item	When the phone rang, John had entered the house.  
	\hyp 	John entered the house before the phone rang (YES).
}

\section*{Acknowledgements}

\ifanon
Anonymized
\else
The research reported in this paper was supported by grant 2014-39 from the
Swedish Research Council, which funds the Centre for Linguistic Theory and
Studies in Probability (CLASP) in the Department of Philosophy, Linguistics,
and Theory of Science at the University of Gothenburg. We are grateful to
our colleagues in CLASP for helpful discussion of some of the ideas presented
here. We also thank anonymous reviewers for their useful comments on an
earlier draft of the paper.
\fi
\bibliographystyle{apalike}

\bibliography{jp,paper2020,NLI}

\end{document}

% Local Variables:
% ispell-local-dictionary: "british"
% End:

% LocalWords:  dowty steedman higginbotham fernando NLI FraCaS GF une
% LocalWords:  aspectual dataset anaphora bernardy grammaticality NL
% LocalWords:  infelicitousness ranta Ljunglof montague english TODO
% LocalWords:  prover chatzikyriakidis Coq's  monadic unger
% LocalWords:  charlow anaphoroi stative activites timespan timespans
% LocalWords:  useCl scandinavian nobel FraCas timepoint Itel unicity
% LocalWords:  Statives contrario th Un lexemes PROG lexicalized lia
% LocalWords:  postfix timepoints FORALL appTime writeUnique FC CCG
% LocalWords:  FraCoq Mineshima Langpro abzianidze premiss UNK Centre
% LocalWords:  Stergios testsuite datasets maccartney compositional
% LocalWords:  Montagovian montagovian coreference neo Davidsonian
% LocalWords:  ogihara bennett iff subinterval APCOM existentials hoc
% LocalWords:  androutsopoulos